%% file: main.tex
\documentclass{article}
\input{z.package}

\input{z.macros}

\title{Fast and Generalizable NeRF Architecture Selection for Satellite Scene Reconstruction}

\author{
Devjyoti Chakraborty$^1$ \and
Zaki Sukma$^{1}$ \and
Rakandhiya D. Rachmanto$^{1}$\and
Kriti Ghosh$^{1}$\and \\
In Kee Kim$^{1}$\and 
Suchendra M. Bhandarkar$^1$ \and 
Lakshmish Ramaswamy$^1$ \and \\
Nancy K. O'Hare$^{2}$ \and
Deepak Mishra$^{2}$
\\
\affiliations
$^1$School of Computing, University of Georgia\\
$^2$Department of Geography, University of Georgia
}

\begin{document}

\maketitle
\thispagestyle{plain}
\pagestyle{plain}

\input{sections/00.abstract}
\input{sections/01.intro}

\input{sections/02.background}
\input{sections/03.approach}
\input{sections/04.eval}

\input{sections/06.device-v2}
\input{sections/07.conclusion}

\bibliographystyle{bibfiles/named}
\bibliography{bibfiles/reference}

\end{document}

%% file: z.package.tex
\usepackage{ijcai26}

\usepackage{times}
\usepackage{soul}
\usepackage{url}
\usepackage[hidelinks]{hyperref}
\usepackage[utf8]{inputenc}
\usepackage[small]{caption}
\usepackage{graphicx}
\usepackage{amsmath}
\usepackage{amsthm}
\usepackage{booktabs}
\usepackage{algorithm}
\usepackage{algorithmic}
\usepackage[switch]{lineno}

\usepackage{algorithm}
\usepackage{algorithmic}
\usepackage{amsmath, amssymb}
\usepackage{tabularx}
\usepackage{booktabs}
\usepackage{pifont} 
\usepackage{multirow}
\usepackage{adjustbox}
\usepackage{makecell}
\usepackage{tikz}
\usepackage[utf8]{inputenc}
\usepackage{newunicodechar}

\usepackage[utf8]{inputenc} 
\usepackage{url}            
\usepackage{booktabs}       
\usepackage{amsfonts}       
\usepackage{nicefrac}       
\usepackage{microtype}      
\usepackage[dvipsnames]{xcolor}
\usepackage{amsmath} 
\usepackage{gensymb}
\usepackage{amsfonts}
\usepackage{enumitem}
\usepackage{graphicx}
\usepackage{xspace}
\usepackage{tikz}
\usepackage{soul}
\usepackage[table]{xcolor} 
\usepackage{booktabs}
\usepackage{colortbl}
\usepackage{multirow}
\usepackage{subcaption}
\usepackage{booktabs}

%% file: z.macros.tex
\newcommand{\eg}{\textit{e.g.}}
\newcommand{\ie}{\textit{i.e.}}

\newcommand{\sysname}{{PreSCAN}\xspace}

\newunicodechar{①}{1}
\newunicodechar{②}{2}
\newunicodechar{③}{3}
\newunicodechar{④}{4}

%% file: sections/00.abstract.tex
\begin{abstract}

Neural Radiance Fields (NeRF) have emerged as a powerful approach for photorealistic 3D reconstruction from multi-view images. 
However, deploying NeRF for satellite imagery remains challenging.
Each scene requires individual training, and optimizing architectures via Neural Architecture Search (NAS) demands hours to days of GPU time. 
While existing approaches focus on architectural improvements, our SHAP analysis reveals that multi-view consistency, rather than model architecture, determines reconstruction quality. 
Based on this insight, we develop \sysname, a predictive framework that estimates NeRF quality prior to training using lightweight geometric and 
photometric descriptors. 
\sysname selects suitable architectures in $<$ 30 seconds with $<$ 1$\,\mathrm{dB}$ prediction error, achieving 1000$\times$ speedup over NAS. 
We further demonstrate \sysname's deployment utility on edge platforms (Jetson Orin), where combining its predictions with offline cost profiling reduces inference power by 26\% and latency by 43\% with minimal quality loss.
Experiments on DFC2019 datasets confirm that \sysname 
generalizes across diverse satellite scenes without retraining.

\end{abstract}

%% file: sections/01.intro.tex
\section{Introduction}

Neural Radiance Fields (NeRF) enable high-quality 3D reconstruction from multi-view images \cite{mildenhall2020nerf}. 
However, satellite imagery applications face severe deployment barriers, as each scene requires individual training and architecture tuning \cite{kilonerf,nerfeval.icpr24}. 
Additionally, the unique characteristics of satellite data, \eg, varying geometry, illumination shifts, and complex RPC camera models, further exacerbate these computational challenges \cite{urbanrad,mari2022sat,vox}. 
These factors make NeRF impractical for large-scale satellite mapping pipelines with thousands of scenes.

Recent advances have focused on accelerating NeRF training through faster optimizers, compact networks, and improved sampling \cite{fastnerf,igp,opnerf,satngp,nerfAcc}. 
While these reduce training time, they do not eliminate the need for scene-specific optimization nor enable cross-scene generalization. 
When Neural Architecture Search (NAS) \cite{nas1000,autotune} is used to optimize NeRF architectures, it demands hours to days of GPU time per scene, {\em prohibitive} for satellite workflows that process thousands of locations/scenes. 
Critically, these approaches prioritize architectural optimization while overlooking the fundamental issue that input view consistency, not model architecture, determines NeRF performance in satellite imagery.
These approaches are built based on a common assumption that architectural optimization is essential for high-quality reconstruction. 
Our work begins by examining the validity of this assumption.

Through extensive SHAP analysis on diverse satellite scenes, we discover that multi-view consistency, rather than model architecture, determines reconstruction quality. 
Specifically, scene-level descriptors (\eg, photometric variance, view angle alignment, geometric coverage) consistently outweigh architectural parameters, including network depth, width, and sampling density. 
This finding suggests that the computational overhead of per-scene NAS can be avoided by predicting reconstruction quality from scene characteristics alone.

Based on this insight, we present {\bf \sysname}, a predictive framework that estimates NeRF reconstruction quality before training, eliminating costly per-scene NAS. 
\sysname jointly encodes architectural parameters and scene-level descriptors (\eg, view alignment, photometric variance, and spatial coverage) to predict reconstruction quality within seconds.
By leveraging both features, \sysname achieves strong generalization with minimal training data without requiring expensive architectural search.

We specifically target satellite imagery, a critical {\em yet largely underexplored} domain for NeRF deployment. 
Our evaluation spans diverse urban environments with varying terrain, lighting, and view configurations. 
Tests on DFC2019 datasets \cite{dfc2019} from Jacksonville (suburban) and Omaha (industrial) demonstrate that 
\sysname maintains prediction accuracy across contrasting scene types without retraining, 
enabling operational remote sensing pipelines.

\sysname achieves a 1000$\times$ speedup over conventional NAS, selecting suitable architectures in $<$ 30 sec. compared to 9+ hours required by NAS on NVIDIA A6000 GPUs.
With $<$ 1$\,\mathrm{dB}$ prediction error, \sysname generalizes across diverse scenes without retraining and remains robust under limited supervision. 
Furthermore, we demonstrate \sysname's applicability to hardware-constrained deployment through a case study on real-world edge computing platforms (\eg, Jetson Orin). By combining \sysname's quality predictions with offline deployment cost profiling, we achieve inference power reductions of 26\% and latency reductions of 43\% with only 0.79$\,\mathrm{dB}$ quality loss, making it practical for edge devices and satellite onboard systems.
These improvements make NeRF-based reconstruction operational for satellite applications where thousands of scenes require processing.

\vspace{1mm}
\noindent {\bf Contributions.} 
Our key contributions are as follows:
\begin{itemize}
    \item[\textbf{(i)}] \textbf{[Scientific Finding]} 
    We demonstrate and provide a quantative validation through SHAP analysis that multi-view consistency, rather than model architecture, determines satellite NeRF reconstruction quality, providing new insight for architecture selection.
    
    \item[\textbf{(ii)}] \textbf{[Framework]} 
    Based on this insight, we introduce \sysname, a predictive framework that estimates NeRF performance before training using lightweight geometric and photometric descriptors, achieving 1000$\times$ speedup over NAS with $<$ 1$\,\mathrm{dB}$ prediction error.
    
    \item[\textbf{(iii)}] \textbf{[Deployment Case Study]} 
    We demonstrate \sysname's practical utility through a hardware-aware deployment case study on edge platforms (Jetson Orin), showing that \sysname's predictions can be combined with offline cost profiling to achieve significant power and latency reductions with minimal quality loss.
\end{itemize}

%% file: sections/02.background.tex
\section{Background and Related Work}
\label{sec:background}

\subsection{NeRF and Its Variants for Satellite Imagery}

\noindent
{\bf NeRF} \cite{mildenhall2020nerf} 
generate photorealistic novel views by learning a continuous volumetric scene function from multi-view color images.
A neural network maps a 3D position $\mathbf{x}$ and viewing direction $\mathbf{d} \in \mathbb{R}^3$ to color and volume density,
$F_\theta(\mathbf{x}, \mathbf{d}) \rightarrow (\mathbf{c}, \sigma)$.

The final rendered color along a ray $r$ is computed as:
\begin{equation}
\mathbf{c}(r) = \sum_{i=1}^{N} T_i \alpha_i \mathbf{c}_i,
\end{equation}
\noindent
with $\alpha_i = 1 - \exp(-\sigma_i \delta_i)$
and $T_i = \prod_{j=1}^{i-1} (1 - \alpha_j)$. 
This formulation allows NeRF to synthesize view-consistent appearance and geometry, 
but it requires per-scene training, limiting its scalability in practice.

Satellite imagery poses unique challenges for NeRF: dynamic elements, long capture intervals, illumination shifts \cite{darbaghshahi2022cloudremoval,wu2023challenges,fassnacht2024remote}, and non-pinhole RPC camera models \cite{grodecki2001ikonos}, all introducing view inconsistencies that often degrade reconstruction quality.

\vspace{2mm}
\noindent
{\bf Shadow NeRF (S-NeRF)} \cite{Derksen_2021_CVPR} 
extends NeRF for satellite imagery by incorporating solar angle $\boldsymbol{\omega}$ into a shadow-aware illumination model:
\begin{equation}
\mathbf{c}(\mathbf{x}, \boldsymbol{\omega}) = \mathbf{c}_a(\mathbf{x}) \cdot \left( s(\mathbf{x}, \boldsymbol{\omega}) + (1 - s(\mathbf{x}, \boldsymbol{\omega})) \cdot \mathbf{a}(\boldsymbol{\omega}) \right)
\label{eq:shadow_coloring}
\end{equation}
where $\mathbf{c}_a(\mathbf{x})$ is the albedo, $s(\cdot)$ is a predicted shadow, and $\mathbf{a}(\cdot)$ is ambient lighting from the sun.
By decoupling geometry and lighting, S-NeRF improves robustness under illumination variation. We adopt S-NeRF as our base model due to its satellite-specific enhancements and computational efficiency. We also use {\bf SatNeRF} \cite{mari2022sat} combined with S-NeRF for testing model invariance. 
SatNeRF defines $F:(\mathbf{x}, \boldsymbol{\omega}, t_j)\!\rightarrow\!(\sigma,\mathbf{c}_a,s,\mathbf{a},\boldsymbol\beta)$.
Relative to S-NeRF, it adds ambient illumination $\mathbf{a}$ and a transient uncertainty term $\boldsymbol{\beta}$, with an additional input dependency of $t_j$, time embeddings attributed to each image.

\subsection{Related Work}

\paragraph{Generalizable Neural Radiance Fields.}
NeRF models typically require scene-specific training and struggle to generalize across environments \cite{yu2020pixelnerf,nerfwild}.
Recent work has proposed generalizable NeRFs \cite{Xu_2023_ICCV,Liu_2024_CVPR,jiang2024gpsnerf,mvsnerf,grf} that improve robustness to sparse inputs, varying illumination, and geometric variation.
However, most methods still require retraining per scene and do not address architectural search overhead.

WaveNeRF \cite{Xu_2023_ICCV} uses wavelet-based decompositions to preserve fine details with few input views, while MVSNeRF \cite{mvsnerf} introduces cost volume aggregation for sparse inputs. Geometry-aware methods like GPSNeRF \cite{jiang2024gpsnerf} and ACA-based NeRFs \cite{Liu_2024_CVPR} fuse geometric, photometric, and sometimes semantic information to improve reconstruction. NeRF Transformer models \cite{attnerf} and Gen-NeRF \cite{mvsnerf} emphasize hardware efficiency or architectural design. Despite these advances, these methods still rely on scene-specific training and do not support fast architecture selection.

\paragraph{Neural Architecture Search (NAS) for NeRF.}
NAS has been applied to reduce NeRF training cost by designing efficient, scene-specific models 
\cite{nas1000,nas-cusr-22,autotune,nair2023nerf,ringnerf,metanerf,evolvnerf}.
NAS-NeRF \cite{nair2023nerf}, for example, uses a generator-inquisitor framework to search over modular MLP architectures.
While such approaches reduce training time, they still require training multiple candidate models per scene, limiting scalability.

In contrast, {\em \sysname avoids per-scene training entirely} by predicting reconstruction quality using lightweight scene-level and architectural descriptors.
This enables fast screening of candidate architectures, reducing NeRF search time from hours to $<$ 30 seconds per scene.

%% file: sections/03.approach.tex
\section{Methodology}\label{sec:method}

\begin{figure*}[t]
  \centering
  \includegraphics[width=1\linewidth]{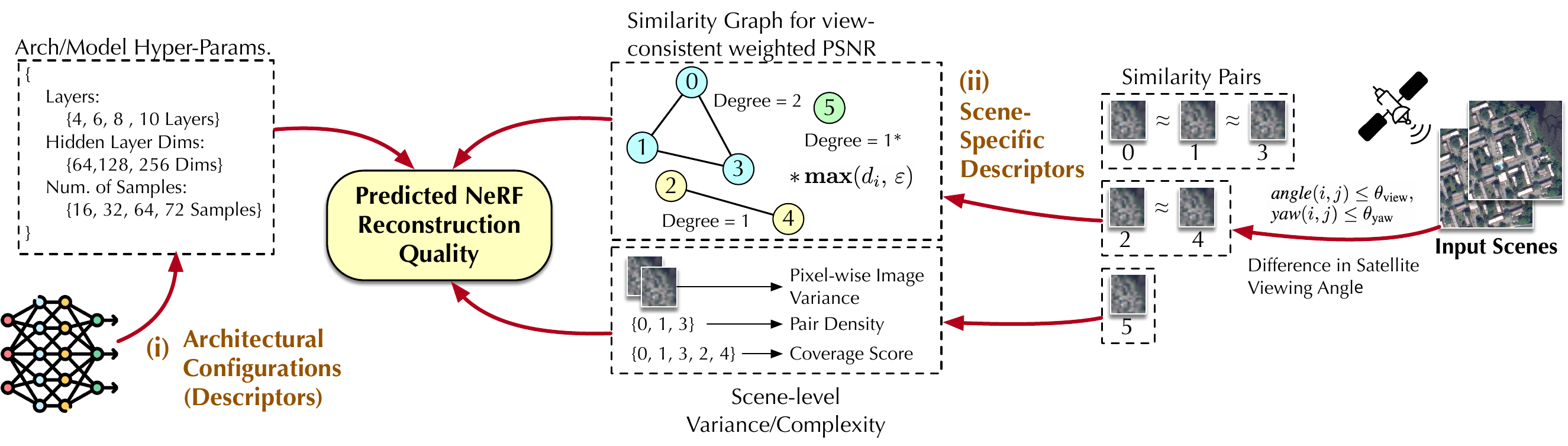}
  \caption{{\bf Overview of the \sysname framework.}
  {\em \sysname predicts suitable NeRF architectures before training by modeling the interaction between architectural parameters and scene descriptors. Using lightweight inputs (\eg, view alignment, sampling patterns), it enables fast ($<$ 30 sec.), generalizable, scalable architecture selection without per-scene retraining.}
  }
  \label{fig:anaswf}
\end{figure*}

\subsection{Scene-Aware NeRF Architecture Prediction}

Our goal is to identify suitable NeRF architectures for satellite scenes {\em prior to any training}, avoiding the costly process of repeatedly training and evaluating models per scene.
\sysname addresses this by estimating NeRF reconstruction performance using a combination of architectural configurations and scene-level descriptors.
This approach eliminates the need for iterative training loops and enables rapid, scalable architecture selection across diverse satellite imagery.

NeRF relies on local ray-based sampling to infer geometry and appearance, making it more sensitive to geometric and photometric consistency than to high-level semantics \cite{Minaee2022ImageSegment,wang2021deep,chen2021review}.
Satellite imagery complicates this process due to shadows, temporal gaps, and transient objects like clouds or vehicles.
To address these challenges, 
\sysname models the interaction between scene-level complexity and architectural design to predict their impact on NeRF reconstruction quality.

Figure~\ref{fig:anaswf} illustrates the \sysname workflow.
The model takes two input types: (i) {\em architectural configurations} (\eg, number of layers, neurons, ray samples), and 
(ii) {\em scene-specific descriptors} that capture photometric variance, ray alignment, and spatial coverage across views.
The two input sets are combined in a lightweight regression model that learns how architectural capacity interacts with scene complexity to influence performance.
Unlike NAS-based approaches that require full model training for each configuration, \sysname enables {\em fast, interpretable architecture screening} with minimal computational overhead.

\subsection{Feature Design and SHAP-based Analysis}\label{subsec:method-shap}

The effectiveness of \sysname relies on input features that accurately reflect the factors influencing NeRF performance.
Unlike traditional vision tasks that benefit from semantic embeddings or high-level abstractions \cite{woo2023convnextv2codesigningscaling,pmlr-v139-tan21a,liu.iccv21}, NeRF reconstruction is more sensitive to pixel-level geometry and color consistency across multiple views.
As a result, NeRF is especially sensitive to variations in viewpoint or appearance, even when the overall scene content appears structurally similar.

We design geometric and photometric descriptors that capture scene consistency across views, including inverse PSNR, view angle and intensity variance, and cosine similarity.
These features quantify multi-view coherence and indicate a scene's suitability for NeRF reconstruction.

To assess feature importance, we perform SHAP analysis \cite{lundberg2017shapley} using multi-view satellite scenes from the DFC2019 dataset \cite{dfc2019}.
We analyze two contrasting regions: Jacksonville (JAX), a suburban area, and Omaha (OMA), an urban industrial zone with distinct terrain and view distributions.

\begin{figure*}[t]
  \centering
  \begin{subfigure}[t]{0.3\linewidth}
    \centering
    \includegraphics[width=\linewidth]{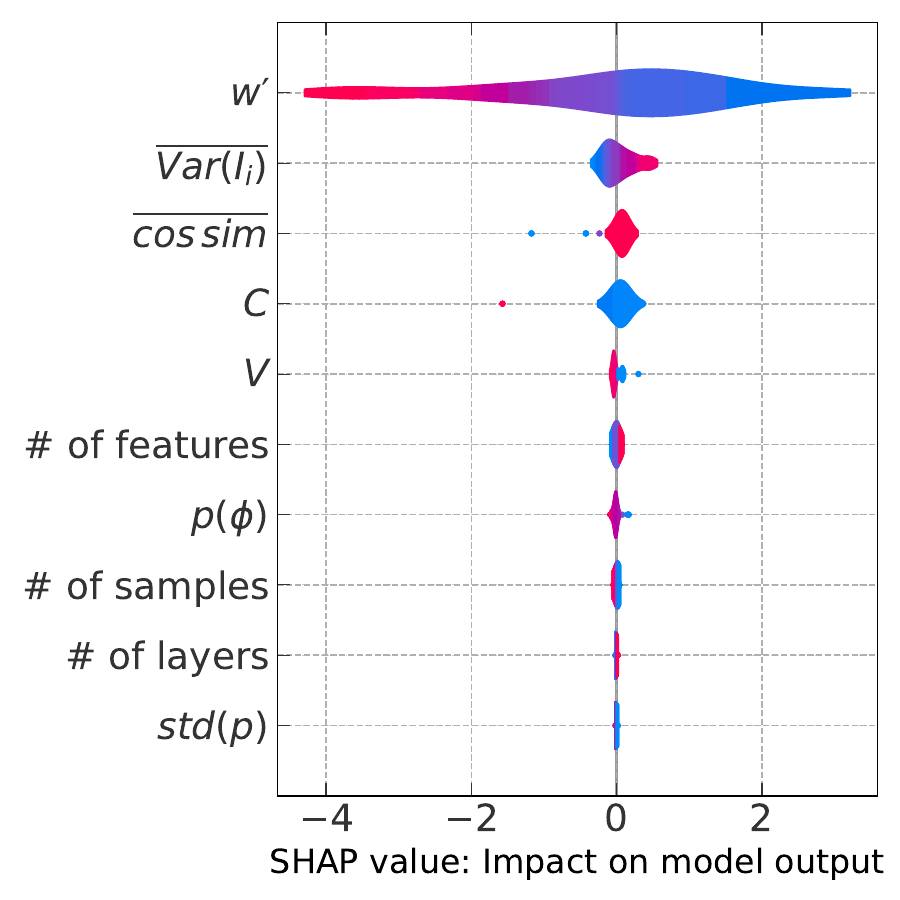}
    \caption{{JAX SHAP values}}
    \label{fig:jax-shap}
  \end{subfigure}
  \begin{subfigure}[t]{0.3\linewidth}
    \centering
    \includegraphics[width=\linewidth]{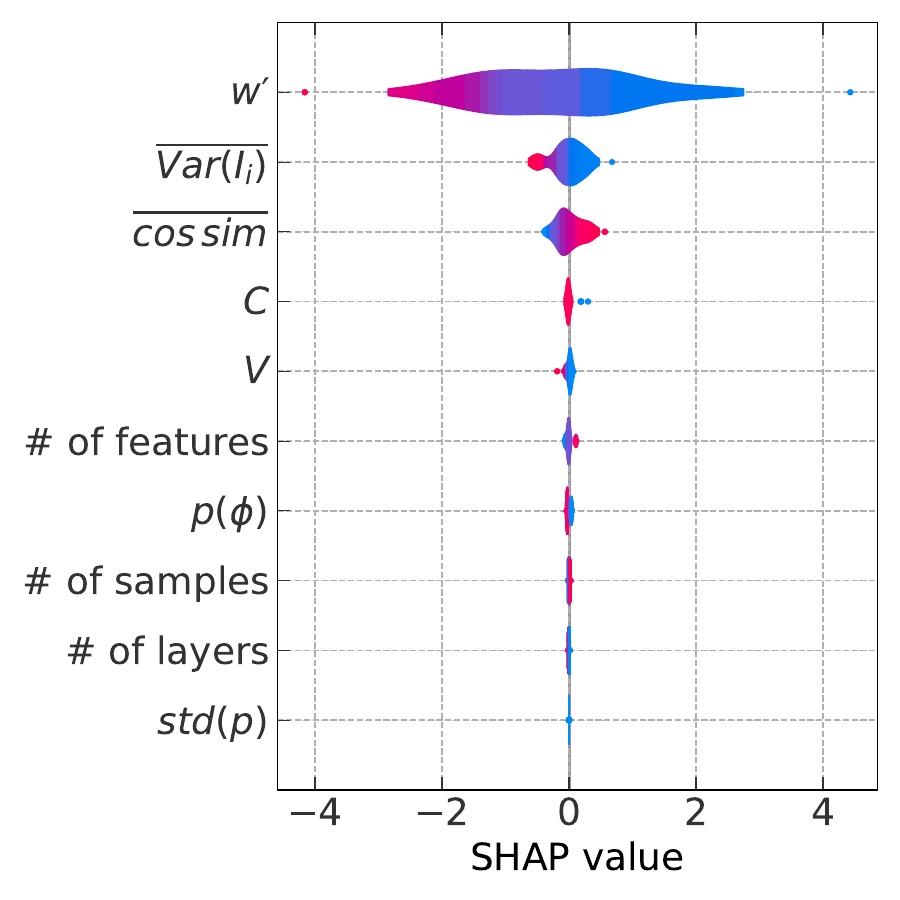}
    \caption{{OMA SHAP values}}
    \label{fig:oma-shap}
  \end{subfigure}
  \begin{subfigure}[t]{0.3\linewidth}
    \centering
    \includegraphics[width=\linewidth]{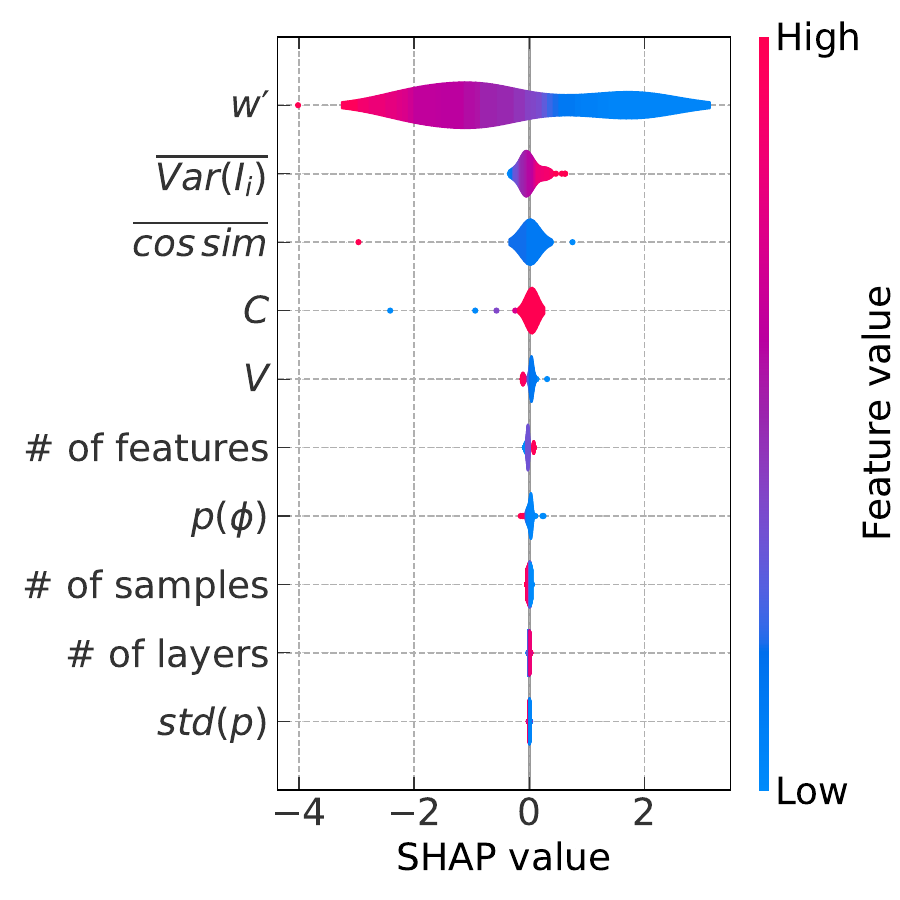}
    \caption{{Mixed (JAX+OMA) SHAP}}
    \label{fig:mixed-shap}
  \end{subfigure}
  \caption{{\bf \bf SHAP value analysis across JAX, OMA, and combined datasets.}
  {\em 
  (a) `JAX scenes': Inverse PSNR ($w'$), photometric variance ($\overline{\text{Var}(I_i)}$), and cosine similarity ($overline{\text{cos sim}}$) are the most influential.
  (b) `OMA scenes': The same top features are consistently identified, despite scene differences.
  (c) `Combined dataset': Feature rankings remain stable, confirming their general importance across datasets.}
  }
  \label{fig:all-shap}
\end{figure*}

As shown in Figure~\ref{fig:jax-shap} and \ref{fig:oma-shap}, SHAP values indicate that inverse PSNR ($w'$), photometric variance ($\overline{\text{Var}(I_i)}$), and view angle cosine similarity ($\overline{\text{cos sim}}$) are the most important predictors across both datasets.
In contrast, architectural parameters, \eg, the number of layers or ray samples, have minimal influence.
This suggests that input consistency plays a more critical role in reconstruction quality than architectural depth or complexity.
We validate this finding by applying SHAP analysis to a combined dataset of JAX and OMA scenes (Figure~\ref{fig:mixed-shap}).
The same three features remain dominant, suggesting that the model captures general patterns relevant to reconstruction, rather than overfitting to dataset-specific artifacts.

In summary, consistency-based features, \eg, {inverse PSNR, photometric variance, and cosine similarity}, are the most influential for predicting NeRF performance. These descriptors generalize across scenes.

Based on this analysis, we hypothesize and validate that \textbf{pixel-level consistency -- {\em both geometric and photometric} -- across nearby views is the key driver of NeRF reconstruction quality} over model specific features.
This hypothesis guides our evaluations, where we assess model generalizability under varying scene conditions.

\subsection{Predictive Feature Representation}\label{subsec:method-predict}
We develop a lightweight feature representation that jointly encodes (i) {\em architectural configurations} and (ii) {\em scene-specific descriptors} (or scene-level geometric and radiometric feastures) to estimate NeRF reconstruction quality  before training.
This allows fast and interpretable performance estimation without requiring rendering or optimization.

\paragraph{(i) Architectural Configurations.}
We use four parameters (descriptors) for each NeRF architecture: 
number of MLP layers ($L$), 
neurons per layer ($f$), 
ray samples per ray ($N_{\text{sample}}$), and 
noise standard deviation ($\sigma_{\text{std}}$).
These parameters capture the model's capacity and inductive biases, influencing its ability to reconstruct geometric and radiometric detail.

\paragraph{(ii) Scene-Specific Descriptors.}
We construct scene-level features capturing view consistency and geometric sampling patterns, both critical for NeRF reconstruction accuracy.

We define valid pairs $(i, j)$ via angular constraints:
\begin{equation}
\mathcal{P} = \{(i, j)\ |\ i < j,\ \text{angle}_{i,j} \leq \theta_\text{view},\ \text{yaw}_{i,j} \leq \theta_\text{yaw}\}, 
\label{eq:valid-pairs}
\end{equation}
where $\theta_\text{view}$ and $\theta_\text{yaw}$ are the pair-wise difference of satellite viewing angle and yaw degree respectively.

The weighted average inverse PSNR $w'$ is defined as:
\begin{equation}
w' = \frac{\sum\limits_{(i,j) \in \mathcal{P}} (d'_i + d'_j) \cdot \text{PSNR}_{ij}^{-1}}{    \sum\limits_{(i,j) \in \mathcal{P}} (d'_i + d'_j)} , 
\label{eq:weight-avg-psnr}
\end{equation}
where $\text{PSNR}_{ij}^{-1} = \frac{1}{\text{PSNR}(I_i, I_j)}$ penalizes dissimilar image pairs, and $d'_i = \max(d_i, \varepsilon)$ assigns a minimum degree to images with no similar neighbors.
The degree weight $d'_i$ favors images with more neighbors ($d_i$ for image $I_i$), with $\varepsilon > 0$ as minimum contribution.

In addition to the weighted inverse PSNR, we compute the pairwise standard deviation of pairwise inverse PSNR values:
\begin{equation}
std(\mathcal{P}) = \sqrt{ \frac{1}{|\mathcal{P}|} \sum_{(i,j)\in \mathcal{P}} \left( \text{PSNR}_{ij}^{-1} - w' \right)^2 }
\end{equation}

To complement the above, we include two additional descriptors.
The {\em pair density} $p(\phi)$ measures the fraction of valid image pairs among all possible view combinations:
\begin{equation}
p(\phi) = \frac{|\mathcal{P}|}{M(M - 1)}, 
\end{equation}
where $M$ is the total number of views.

The \emph{coverage score} $\text{C}$ quantifies the proportion of input images that participate in at least one valid pair:
\begin{equation}
\text{C} = \frac{1}{M} \sum_{i=1}^M c_i, \quad \text{where } c_i = 1 \text{ if } I_i \in \mathcal{P}
\end{equation}

We compute the \emph{mean cosine similarity} $\overline{\text{cos sim}}$ between view direction vectors to quantify angular alignment across multi-view observations. Higher values indicate strong directional coherence, which benefits volumetric integration.
\begin{equation}
\overline{\text{cos sim}} = \frac{1}{|\mathcal{P}|} \sum_{(i,j)\in \mathcal{P}} \frac{\mathbf{d}_i \cdot \mathbf{d}_j}{|\mathbf{d}_i||\mathbf{d}_j|},
\end{equation}
where $\mathbf{d}_i$ and $\mathbf{d}_j$ are the unit view direction vectors for cameras $i$ and $j$, and $\mathcal{P}$ is the set of valid image pairs.

The \emph{image variance} $\overline{\text{Var}(I_i)}$ captures photometric diversity across a scene, including shadows, occlusions, or surface variation that challenge radiometric modeling.
\begin{equation}
\overline{\text{Var}(I_i)} = \frac{1}{M} \sum_{i=1}^{M} \text{Var}(I_i), 
\label{eq:var_img}
\end{equation}
\noindent
where $\text{Var}(I_i)$ is the pixel-level variance of image $I_i$.

Additionally, we include view-angle statistics (mean, max, std) to capture the overall camera configuration.

\paragraph{Prediction Model.} 
We concatenate architectural and scene-level descriptors into a unified feature vector and use a parametric performance estimator to predict reconstruction performance:
\begin{equation}
\text{PSNR}_{\text{pred}} = \boldsymbol{\beta}_{\text{arch}}^\top \mathbf{X}_{\text{arch}} + \boldsymbol{\beta}_{\text{scene}}^\top \mathbf{X}_{\text{scene}} + \beta_0,
\label{eq:prescan-model}
\end{equation}
where $\mathbf{X}_{\text{arch}}$ represents architectural configurations $(L, f, N_{\text{sample}}, \sigma_{\text{std}})$, and $\mathbf{X}_{\text{scene}}$ 
denotes scene-level features 
$(w', \text{std}(p), p(\phi), C, \overline{\text{cos sim}}, \overline{\text{Var}(I_i)})$. 
The vectors $\boldsymbol{\beta}_{\text{arch}}$ and $\boldsymbol{\beta}_{\text{scene}}$ are the learned weights for each feature type, and $\beta_0$ is a scalar bias term.

We deliberately choose  a parametric shallow performance estimator over more complex models (\eg, deep neural networks, gradient boosting) for three reasons: 
(1) our SHAP analysis reveals approximately linear relationships between scene descriptors and reconstruction quality, 
(2) linear models provide direct interpretability of feature contributions, and 
(3) they generalize well under limited training data without overfitting. Our evaluation confirms that this simple model achieves $<$ 1$\,\mathrm{dB}$ prediction error, demonstrating that additional model complexity is unnecessary.

%% file: sections/04.eval.tex
\section{Evaluation}\label{sec:eval}
\begin{table}[t]
\centering
\caption{{\bf PSNR statistics across two datasets. }{\em  The results show diversity in reconstruction difficulty and complexity.}}
{\resizebox{0.65\columnwidth}!
{
  \begin{tabular}{c|c|c|c}
  \toprule
  \textbf{Dataset} & \begin{tabular}[c]{@{}c@{}}\textbf{PSNR}\\ {(Average)}\end{tabular} & \begin{tabular}[c]{@{}c@{}}\textbf{PSNR}\\ {(Max)}\end{tabular}  & \begin{tabular}[c]{@{}c@{}}\textbf{PSNR}\\ {(Min)}\end{tabular}  \\
  \midrule
  JAX & 19.63 & 23.03 & 14.16 \\
  OMA        & 17.18 & 23.06 & 12.21 \\
  \bottomrule
  \end{tabular}
}
}
\label{tab:psnr_summary}
\end{table}

\subsection{Evaluation Setup}

We evaluate \sysname on 73 satellite scenes (33 JAX, 40 OMA) from DFC2019 \cite{dfc2019} using S-NeRF \cite{Derksen_2021_CVPR} as the base model, extracting six scene descriptors (discussed in the previous section) per scene.  
NAS uses an A6000, while \sysname runs on a Tesla T4 (a weaker GPU than the A6000).

\subsection{\sysname Model Construction}
\label{subsec:dataset-model}

Running NAS on S-NeRF across 73 scenes yields 3650 scene-architecture pairs (50 top-performing architectures per scene) for training our prediction model (Eq.\ref{eq:prescan-model}). 
The NAS search space is defined over architectural parameters known to significantly impact NeRF performance.
\begin{itemize}
\item {\bf Number of MLP layers} ($L$): Chosen from $\{4, 6, 8, 10\}$ to span shallow to deeper networks.
\item {\bf Hidden layer dimensions} ($f$): Selected from $\{64, 128, 256\}$ to control model capacity.
\item {\bf Samples per ray} ($N_{\text{sample}}$): Drawn from $\{16, 32, 64, 72\}$ to vary volumetric resolution.
\item {\bf Sampling noise standard deviation} ($\sigma_{\text{std}}$): Uniformly sampled from $\{0.0, 0.12, 0.14\}$ for noise robustnesses.
\end{itemize}

\begin{table}[t]
  \centering
  \caption{{\bf Comparison of NAS (on A6000 GPU) and \sysname\ (on Tesla T4, weaker GPU) prediction time.} {\em \sysname achieves $>$ 1000$\times$ speedup over NAS with comparable accuracy, enabling sub-minute ($<$ 30 sec.) prediction per scene.}}
  \vspace{0.5ex}
    \begin{subtable}[t]{0.35\linewidth}
    \centering
    \caption{Average}
      \resizebox{0.95\columnwidth}{!}{
      \begin{tabular}{lc}
        \toprule
        {Method} & {Time} \\
        \midrule
        NAS      & 9.7 hrs \\
        {\bf \sysname}  & {\bf 28.3s} \\
        \bottomrule
      \end{tabular}
      }
      \end{subtable}
      \hfill
      \begin{subtable}[t]{0.6\linewidth}
      \centering
      \caption{Three Example Scenes}
      \resizebox{0.9\columnwidth}{!}{
      \begin{tabular}{lcc}
        \toprule
        {Scene} & {NAS} & {\bf \sysname} \\
        \midrule
        JAX-467 & 10.1 hrs & {\bf 28.6s} \\
        JAX-258 & 9.6 hrs  & {\bf 28.9s} \\
        OMA-353 & 9.9 hrs  & {\bf 28.4s} \\
        \bottomrule
      \end{tabular}
      }
    \end{subtable}
  \label{tab:nas-comparison}
\end{table}

The hyperparameter combinations yield 144 possible S-NeRF configurations. We use TPE \cite{watanabe2023tpe} to select the top-50 architectures per scene via NAS, requiring $\approx$10 hours per scene on an A6000 GPU. Table \ref{tab:psnr_summary} shows PSNR variations across two datasets, confirming scene diversity.

We train on 64 scenes (3200 pairs) using ordinary least squares, which converges in under one minute. Testing on 9 remaining scenes (four JAX, five OMA) demonstrates that \sysname achieves $>$ 1000$\times$ speedup over NAS while maintaining accuracy (Table~\ref{tab:nas-comparison}).

\begin{table}[t]
\centering
\caption{
{\bf Comparison of actual vs. predicted average PSNR for scenes.}
{\em We report MAE (absolute PSNR gap, $\Delta$PSNR = $|$Actual $-$ Predicted$|$).
Bold cells indicate low MAE (best predictions), and values marked as (worst) denote high error.}
}
\resizebox{0.95\columnwidth}{!}{
  \begin{tabular}{l|ccccc}
  \toprule
  \textbf{Scene} 
    & \begin{tabular}[c]{@{}c@{}}\textbf{Actual}\\ \textbf{PSNR}\end{tabular} 
    & \begin{tabular}[c]{@{}c@{}}\textbf{Predict.}\\ \textbf{PSNR}\end{tabular} 
    & \begin{tabular}[c]{@{}c@{}}\textbf{MAE}\\ \textbf{($\boldsymbol{\Delta}$PSNR)}\end{tabular}  
    & \begin{tabular}[c]{@{}c@{}}\textbf{Std.}\\ \textbf{Error}\end{tabular}  
    & \textbf{Bias}  \\
  \midrule
  JAX-467 & 20.22 & 20.74 & 0.52 & 0.71 & +0.52 \\
  JAX-474 & \textbf{20.15} & \textbf{20.57} & \textbf{0.42 (best)} & \textbf{0.31} & \textbf{+0.42} \\
  JAX-505 & 19.14 & 19.70 & 0.56 & 0.74 & +0.56 \\
  JAX-559 & \textbf{19.06} & \textbf{19.95} & \textbf{0.89 (worst)} & \textbf{0.62} & \textbf{+0.89} \\
  \midrule
  OMA-258 & \textbf{19.10} & \textbf{19.04} & \textbf{0.07 (best)} & \textbf{0.13} & \textbf{-0.07} \\
  OMA-287 & 15.88 & 16.14 & 0.27 & 0.21 & +0.27 \\
  OMA-315 & 17.12 & 16.64 & 0.48 & 0.17 & -0.48 \\
  OMA-331 & \textbf{15.59} & \textbf{16.49} & \textbf{0.89 (worst)} & \textbf{0.20} & \textbf{+0.89} \\
  OMA-353 & 16.74 & 17.26 & 0.52 & 0.13 & +0.52 \\
  \bottomrule
  \end{tabular}
}
\label{tab:psnr_comparison}
\end{table}

\begin{table}[t]
\centering
\caption{{\bf Scene-wise PSNR Generalization Comparison: Sparse Training (45 scenes)}.
{\em Bold cells indicate low MAE (best), and values marked as (worst) denote high MAE error.}}
\resizebox{0.95\columnwidth}{!}{
  \begin{tabular}{l|cccc}
    \toprule
    \multicolumn{5}{c}{\textbf{JAX (Unseen Scenes)}} \\
    \midrule
    \textbf{Scene} & \textbf{Act. PSNR} & \textbf{Pred. PSNR} & \textbf{MAE} & \textbf{Bias} \\
    \midrule
    JAX-276 & 19.88 & 19.24 & 0.64 & -0.64 \\
    JAX-280 & 18.95 & 18.70 & 0.25 & -0.25 \\
    JAX-314 & \textbf{18.53} & \textbf{18.60} & \textbf{0.08 (best)} & \textbf{+0.08} \\
    JAX-359 & \textbf{17.94} & \textbf{18.77} & \textbf{0.83 (worst)} & \textbf{+0.83} \\
    JAX-412 & 18.46 & 19.11 & 0.65 & +0.65 \\
    JAX-416 & 19.40 & 19.28 & 0.13 & -0.13 \\
    \midrule
    \multicolumn{5}{c}{\textbf{OMA (Unseen Scenes)}} \\
    \midrule
    \textbf{Scene} & \textbf{Actual} & \textbf{Pred.} & \textbf{MAE} & \textbf{Bias} \\
    \midrule
    OMA-142 & 20.19 & 19.93 & 0.26 & -0.26 \\
    OMA-212 & 16.54 & 15.88 & 0.66 & -0.66 \\
    OMA-244 & \textbf{19.31} & \textbf{17.46} & \textbf{1.85 (worst)} & \textbf{-1.85} \\
    OMA-247 & 15.55 & 16.06 & 0.51 & +0.51 \\
    OMA-276 & 15.49 & 16.84 & 1.34 & +1.34 \\
    OMA-278 & 15.80 & 16.30 & 0.50 & +0.50 \\
    OMA-355 & 17.25 & 17.21 & 0.04 & -0.04 \\
    OMA-376 & 18.57 & 18.22 & 0.35 & -0.35 \\
    OMA-389 & \textbf{17.88} & \textbf{17.89} & \textbf{0.01 (best)} & \textbf{+0.01} \\
    \bottomrule
  \end{tabular}
}
\label{tab:jax_oma_sparse}
\end{table}

\begin{figure*}[htbp]
  \centering
  \includegraphics[width=0.95\linewidth]{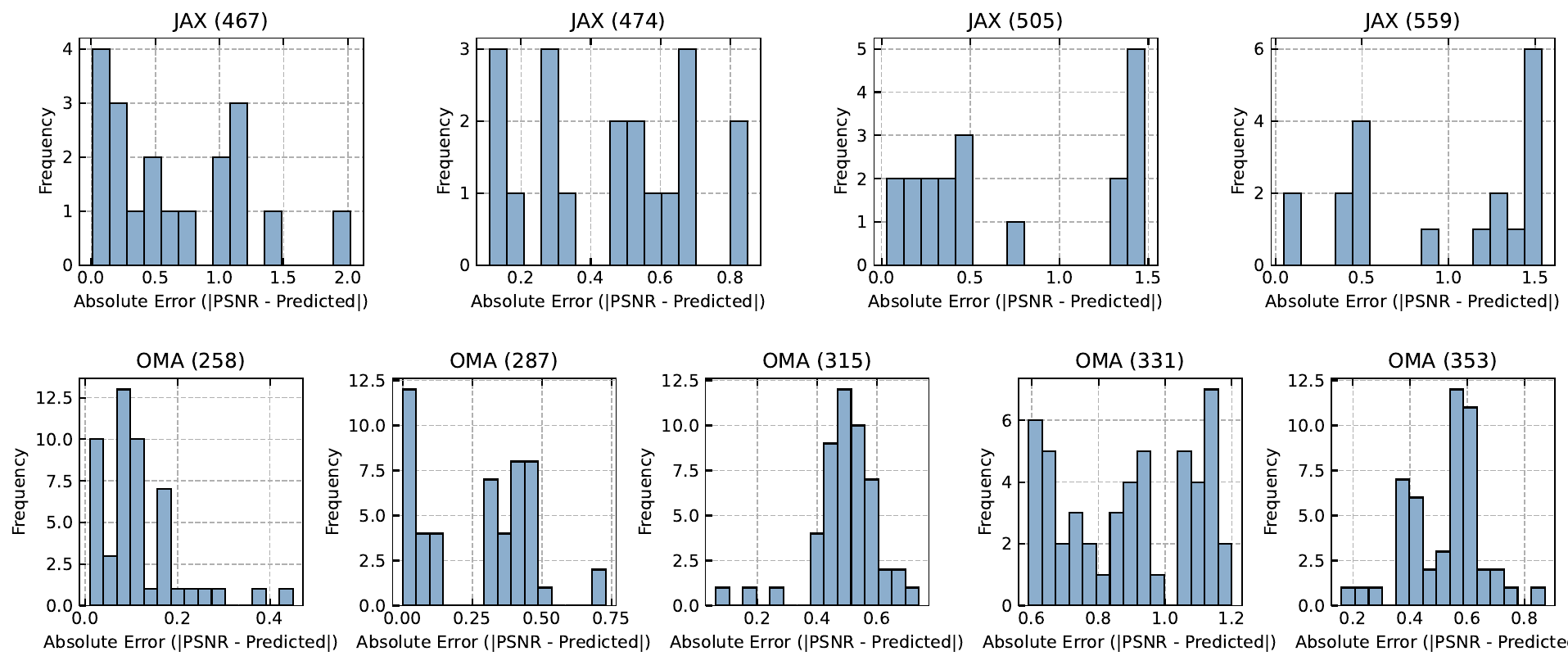}
  \caption{{\bf MAE ($\Delta$PSNR) distributions for different NeRF architectures across scenes.} {\em Each subplot shows prediction 
consistency per scene, with most errors $<$ 1$\,\mathrm{dB}$ dB and many 
clustering $<$ 0.5$\,\mathrm{dB}$, demonstrating \sysname's robust 
generalization.}}
  \label{fig:anaspredstan}
\end{figure*}

\subsection{Predictive Performance of PSNR Estimation}

We evaluate \sysname on nine test scenes (4 JAX, 5 OMA) with varying reconstruction complexity. 
Table~\ref{tab:psnr_comparison} reports predicted vs. actual PSNR and error metrics, \eg, each scene's MAE ($\Delta$PSNR), bias, and standard deviation of errors.

\sysname achieves MAE of 0.06 -- 0.89$\,\mathrm{dB}$ in PSNR across nine test scenes, with most predictions within 0.5$\,\mathrm{dB}$ of ground truth. 
JAX scenes exhibit slight overprediction (positive bias), while OMA scenes vary more: OMA-258 shows excellent accuracy (0.07$\,\mathrm{dB}$ MAE), whereas JAX-559 and OMA-331 reach the upper error bound ($\approx$0.89$\,\mathrm{dB}$). 
All errors remain practical for deployment and show \sysname's robust and reliable NeRF prediction ability.

Our results reveal a critical insight: multi-view consistency among nearby viewpoints drives PSNR prediction accuracy more than architectural choices. 
Figure~\ref{fig:anaspredstan} validates this finding as \sysname maintains $<$ 1$\,\mathrm{dB}$ error across all architectures and scenes, with most achieving $<$ 0.5$\,\mathrm{dB}$ precision. 
Even challenging scenes (JAX-559, OMA-331) remain within bounds, while well-aligned scenes (OMA-258) achieve 0.07$\,\mathrm{dB}$ MAE. 
This validates \sysname's consistent performance despite scene variations.

\begin{figure*}[t]
  \centering
  \includegraphics[width=0.9\linewidth]{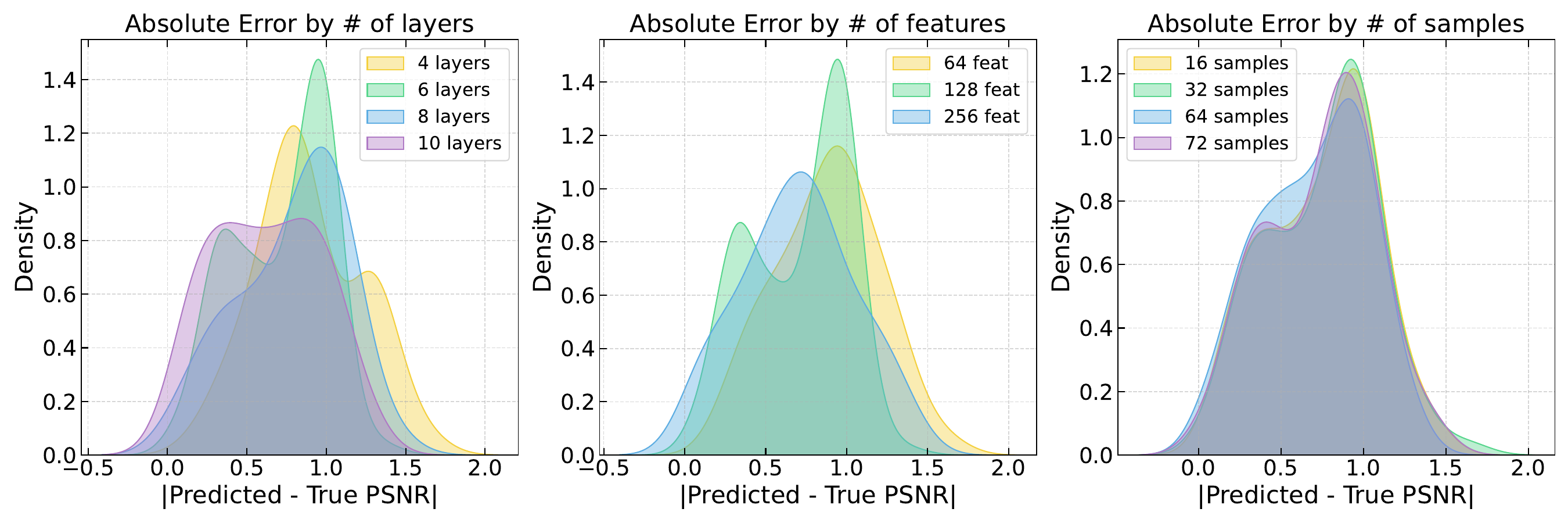}
  \caption{{\bf Distribution of absolute prediction errors ($\Delta$PSNR) across various architectural parameters.}
{\em (a) `Layers': NeRF models with 6 and 8-layer yield tighter, more stable error distributions, while models with 10-layer shows higher spread and occasional outliers.}
{\em (b) `Feature Dimensionality': 128-feature models produce the most consistent predictions.}
{\em (c) `Number of Samples': Increasing the number of samples $>$ 32 offers diminishing returns, with all settings clustering near 1$\,\mathrm{dB}$ error.}}
  \label{fig:kde}
\end{figure*}

\subsection{Architectural Error Distributions }

Beyond average performance metrics, we analyze how architectural parameters affect prediction accuracy by examining $\Delta$PSNR distributions across key design choices, 
including layers, features, and sampling density (Figure~\ref{fig:kde}).

Figure~\ref{fig:kde} reveals hidden architectural trends. 
Models with 128 features show tighter error distributions than 64-feature variants despite similar means, indicating superior robustness. 
10-layer networks exhibit wider error spreads, suggesting overfitting, while sampling beyond 32 rays yields diminishing returns. 
These findings validate that model capacity outweighs sampling density for satellite NeRF performance.

These architectural trends demonstrate \sysname's ability to capture fine-grained performance patterns beyond scene-level statistics. 
The distinct error profiles for different depths and capacities confirm that \sysname generalizes across both scene diversity and architectural variations, essential for practical NAS replacement. 
Such predictable error distributions can enable uncertainty- or confidence-aware architecture selection in future deployment scenarios.

\subsection{Sparse Input Model Training}

To assess generalization under limited supervision, we retrain \sysname using only 45 scenes. 
The {\em linear relationship between descriptors and PSNR should enable strong performance even in data-scarce settings}, a critical requirement for new geographic regions. 
Table~\ref{tab:jax_oma_sparse} validates this hypothesis. 
\sysname maintains $<$ 1$\,\mathrm{dB}$ errors on most unseen scenes, demonstrating robustness in sparse training scenarios.

Results confirm robust generalization across scene types. 
Most predictions achieve excellent accuracy: 
JAX-314 (0.08$\,\mathrm{dB}$) and 
OMA-142 (0.26$\,\mathrm{dB}$) show particularly low errors. 
Even challenging cases like 
JAX-359 (0.83$\,\mathrm{dB}$) and OMA-244 (1.85$\,\mathrm{dB}$, due to extreme viewing geometry) 
remain practical. 
Moreover, we observe that prediction errors correlate strongly with actual scene difficulty. \ie, higher errors reliably indicate problematic inputs requiring special handling. This property enables \sysname to serve dual roles in operational pipelines: 1) selecting appropriate architectures and 2) flagging scenes for quality review.

\subsection{Model Invariancy}

To test our approach against model invariance, we introduce an additional branch that models architecture-specific effects as
$\boldsymbol{\beta}^{\top}\mathbf{X}_{\text{model}}$ where $\mathbf{X}_{\text{model}}$ denotes the model variant used used. In our training database, we include S-NeRF and SatNeRF as model choices.

\begin{table}[t]
\centering
\caption{
{\bf Comparison of MAE (Actual $-$ Predicted PSNR) across NeRF variants.}
{\em We report absolute PSNR error (Predicted - Actual) for S-NeRF and SatNeRF.}
}
\resizebox{0.8\columnwidth}{!}{
  \begin{tabular}{l|cc}
  \toprule
  \textbf{Scene} 
    & \begin{tabular}[c]{@{}c@{}}\textbf{S-NeRF MAE}\\ {($\boldsymbol{\Delta}$PSNR)}\end{tabular} 
    & \begin{tabular}[c]{@{}c@{}}\textbf{SatNeRF MAE}\\ {($\boldsymbol{\Delta}$PSNR)}\end{tabular} \\
  \midrule
  JAX-467 & 0.79 & 0.82 \\
  JAX-474 & 0.55 & 0.77 \\
  JAX-505 & 0.61 & 0.91 \\
  JAX-559 & 0.03 & 0.57 \\
  \midrule
  OMA-258 & 0.19 & 0.02 \\
  OMA-287 & 0.37 & 0.56 \\
  OMA-315 & 0.31 & 0.41 \\
  OMA-331 & 0.72 & 0.49 \\
  OMA-353 & 0.88 & 0.17 \\
  \bottomrule
  \end{tabular}
}
\label{tab:nerf_variant_mae}
\end{table}

As shown in Table~\ref{tab:nerf_variant_mae}, the prediction error remains consistently below 1$\,\mathrm{dB}$ across both S-NeRF and SatNeRF variants, indicating that our approach captures scene-driven reconstruction difficulty rather than model-specific biases, enabling Nerf variant-agnostic performance estimation.

%% file: sections/06.device-v2.tex
\begin{figure*}[t]
    \centering
    \includegraphics[width=\linewidth]{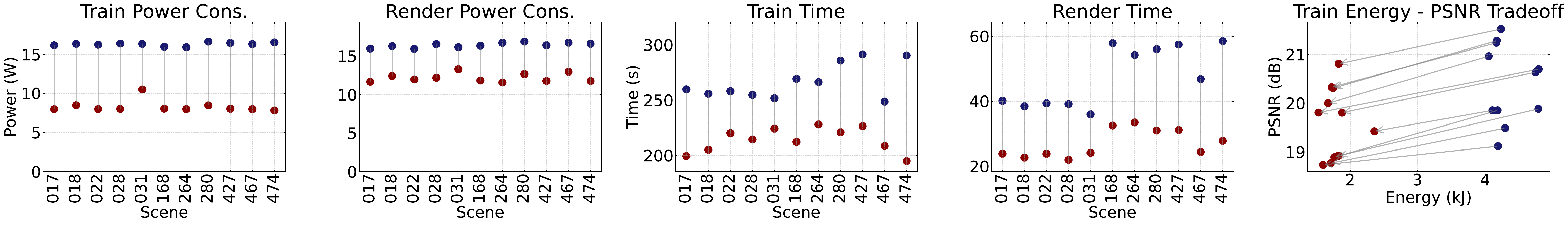}
    \caption{\small {\bf Comparison of power consumption and time between the most accurate model and the cost-efficient model on the edge device (Blue indicates baseline, red indicates \sysname chosen optimal).} 
    {\em (a) Training power consumption, (b) Rendering power consumption, 
    (c) Training time, (d) Rendering time, (e) Energy-PSNR tradeoff. 
    \sysname with hardware-aware architecture selection achieves 
    significant efficiency gains with minimal quality loss.}}
    \label{fig:deployment_cost_comparison}
\end{figure*}

\section{Case Study: Hardware-Aware Deployment}\label{sec:hadp}

\sysname enables fast architecture selection by predicting reconstruction quality prior to training. 
However, real-world deployment on satellite onboard systems, UAVs, and edge devices imposes strict power and latency budgets that often preclude selecting the highest-accuracy (typically largest) architecture. 

To demonstrate \sysname's practical value in resource-constrained settings, we present a deployment case study on edge platforms. 
Specifically, we show that \sysname's scene-aware quality predictions can be directly combined with offline hardware cost profiles to enable cost-efficient architecture selection without additional model training or modification.

\subsection{Offline Deployment Cost Profiling}

Deployment costs (latency and power consumption) are primarily driven by architectural parameters (\eg, network depth, width, sampling density) rather than scene content. 
Unlike reconstruction quality, which varies by scene, the computational cost for a fixed architecture remains consistent. 
This independence allows us to decouple cost estimation from scene-specific quality prediction.

Leveraging this, we construct a deployment cost database by profiling all candidate NeRF architectures offline on target edge platforms, specifically Jetson Orin Nano and Orin AGX. 
For each architecture, we measure four key metrics: training power, rendering power, training time, and rendering time. 
This one-time profiling creates a lookup table that complements \sysname's scene-aware quality predictions, enabling instant cost-benefit analysis for any new satellite scene.

\subsection{Hardware-Aware Architecture Selection}

We aim to identify the architecture that minimizes energy and latency while meeting a minimum quality requirement. 
Given a target device and a quality threshold $PSNR_{\min}$, the selection process follows a simple three-step logic:
\begin{enumerate}
    \item \textbf{Predict Quality:} \sysname estimates the PSNR for all candidate architectures for the specific scene.
    \item \textbf{Filter:} Architectures failing to meet $PSNR_{\min}$ are discarded.
    \item \textbf{Optimize Cost:} Among the remaining valid candidates, we query the deployment cost database to select the architecture with the lowest energy consumption or latency.
\end{enumerate}
This approach bypasses iterative training or hardware-in-the-loop testing during operation.
Please note that this selection pipeline requires no modification to \sysname itself; it operates entirely on \sysname's existing quality predictions combined with precomputed cost profiles.

\subsection{Empirical Efficiency Gains}

We evaluate this hardware-aware strategy by comparing it against an unconstrained baseline (Best Accuracy) that selects the architecture with the highest predicted PSNR. 
For this evaluation, we applied a strict quality budget relative to the maximum achievable accuracy for a set of test scenes.

Figure~\ref{fig:deployment_cost_comparison} visualizes the trade-offs between deployment cost and reconstruction performance. 
The results indicate that accepting a marginal difference in reconstruction quality allows for the selection of significantly more efficient architectures. 
As summarized in Table~\ref{tab:device-metrics-summary}, this approach yields substantial reductions in both power consumption and latency for training and rendering, with a negligible impact on the final PSNR. 
These findings confirm that \sysname bridges the gap between high-fidelity NeRF reconstruction and the strict resource limitations of satellite onboard systems or edge platforms.

\begin{table}[t]
\caption{\bf Average percentage of benefit from hardware-aware architecture selection with the tradeoff in accuracy.}
  \centering
  \resizebox{0.7\columnwidth}{!}{
  \begin{tabular}{lc}
    \toprule
    PSNR Diff & 0.79$\,\mathrm{dB}$  \\
    MAE Diff & 0.4 m \\
    \midrule
    Power Consumption -- Train & 49\%  \\
    Power Consumption -- Render & 26\% \\
    Time Reduction -- Train & 19\%  \\
    Time Reduction -- Render & 42\% \\
    \bottomrule
  \end{tabular}}
  
  \label{tab:device-metrics-summary}
\end{table}

%% file: sections/07.conclusion.tex
\section{Conclusion}

We present \sysname, a scene-aware NeRF architecture selection framework that predicts reconstruction quality without full model training or NAS.
By combining architectural and photometric–geometric descriptors, \sysname enables 1000$\times$ faster architecture screening with $<$ 1$\,\mathrm{dB}$ MAE across diverse satellite scenes.
SHAP analysis shows that view consistency is more predictive than architectural complexity, and the model reliably generalizes under sparse supervision.
We further demonstrate \sysname's practical applicability through a hardware-aware deployment case study on edge platforms, showing that its predictions can be combined with offline cost profiling to achieve significant reductions in power and latency with minimal quality loss. 
Our future work includes expanding to other models (\cite{spsnerf-arxiv-23}, \cite{Eonerf}), evaluating on more diverse hardware platforms, and a larger dataset to further validate \sysname's generalization for broader operational deployment.